\documentclass[letterpaper]{article}
\usepackage{aaai2027}
\usepackage[hyphens]{url}
\usepackage{graphicx}
\usepackage{natbib}
\usepackage{caption}
\usepackage{xcolor}
\usepackage{pifont}
\usepackage{placeins}
\usepackage{pifont}
\usepackage{array}
\newcommand{\cmark}{\textcolor{green!50!black}{\ding{51}}}
\newcommand{\xmark}{\textcolor{red!70!black}{\ding{55}}}
\newcommand{\pmark}{\textcolor{orange!85!black}{\ensuremath{\boldsymbol{\sim}}}}
\usepackage{tabularx}
\newcolumntype{L}[1]{>{\raggedright\arraybackslash}m{#1}}
\newcolumntype{C}[1]{>{\centering\arraybackslash}m{#1}}
\usepackage{amsmath}
\usepackage{amssymb}
\usepackage{booktabs}
\usepackage{multirow}
\usepackage{array}
\usepackage{siunitx}
\usepackage{tcolorbox}
\usepackage{algorithm}
\usepackage{algorithmic}
\usepackage{newfloat}
\usepackage{listings}
\usepackage{amssymb}
\usepackage{enumitem}
\newcolumntype{C}[1]{>{\centering\arraybackslash}p{#1}}

\title{
\raisebox{-0.25em}{\includegraphics[height=1.40em]{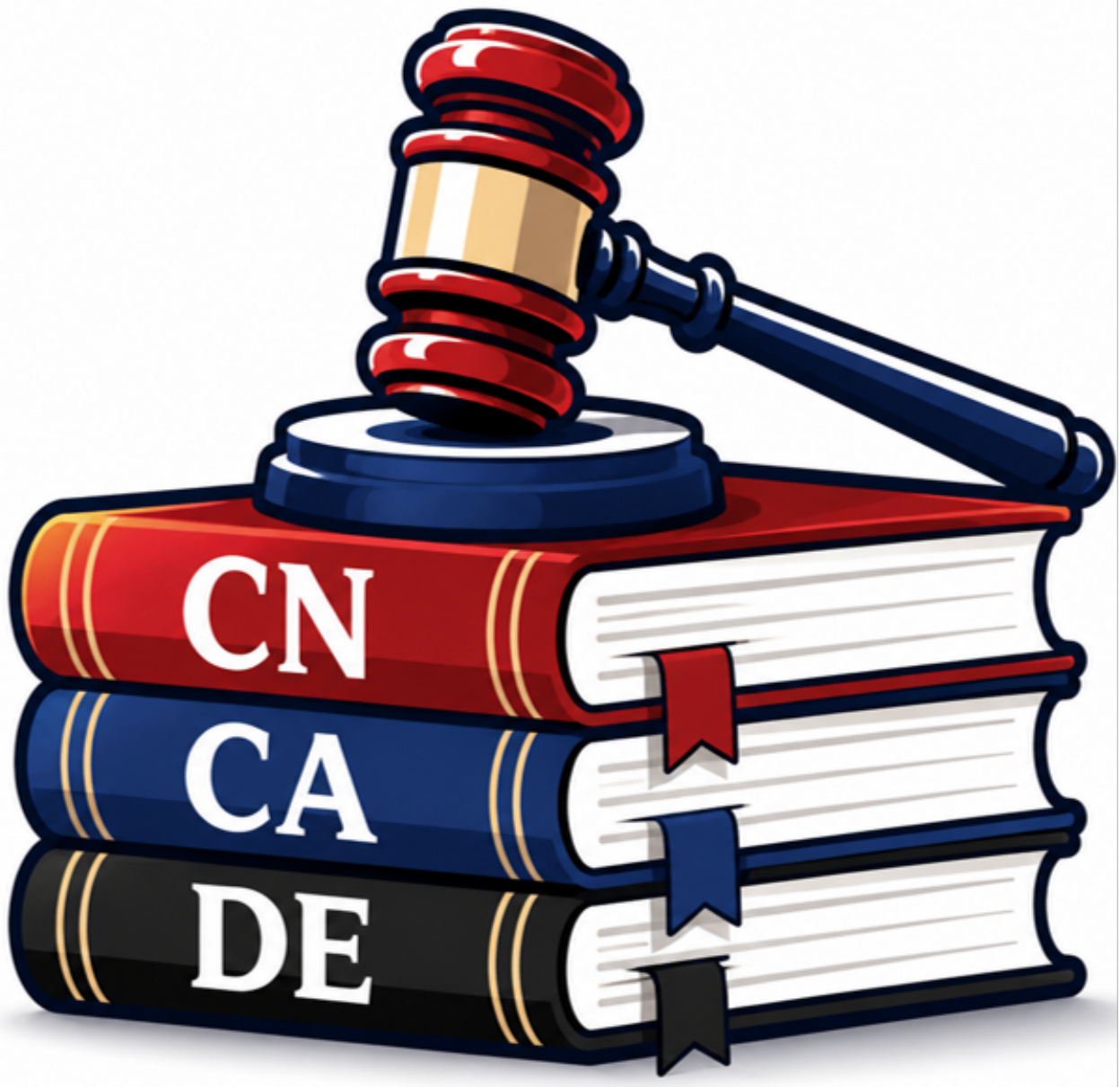}}
\hspace{0.25em}
\textsc{CrossLex}: A Source-Grounded Benchmark for Cross-Jurisdictional Legal Reasoning in Large Language Models
}

\author{
    Xiaocui Yang\textsuperscript{\rm 1},
    Xican Tan\textsuperscript{\rm 2},
    Shoujie Chen\textsuperscript{\rm 1},
    Shihan Xiao\textsuperscript{\rm 3},
    Keke Tong\textsuperscript{\rm 4},
    Xinyu Zhou\textsuperscript{\rm 2}\corresponding
}
\affiliations{
    \textsuperscript{\rm 1}School of Computer Science and Engineering, Northeastern University, Shenyang, China\\
    \textsuperscript{\rm 2}Institute of Intelligent Computing, University of Electronic Science and Technology of China, Chengdu, China\\
    \textsuperscript{\rm 3}School of Civil and Commercial Law, Southwest University of Political Science and Law, Chongqing, China\\
    \textsuperscript{\rm 4}Chongqing Branch of China Unicom, Chongqing, China\\
    yangxiaocui@cse.neu.edu.cn, chenshoujie2003@163.com,
    2022214975@stu.cqupt.edu.cn,\\
    202305255416@smail.xtu.edu.cn,
    2022214975@stu.cqupt.edu.cn, zhxy@uestc.edu.cn
}

\begin{document}
\maketitle
\begin{abstract}
Legal reasoning is inherently jurisdiction-dependent, i.e., the same facts can call for different legal rules and yield different conclusions across legal systems. Yet existing benchmarks rarely evaluate whether Large Language Models (LLMs) can recognize such jurisdiction-specific variation, especially when identical fact patterns lead to divergent legal outcomes.
We introduce \textsc{CrossLex}\footnote{Relevant resources will be released upon paper acceptance.}, a same-fact, legal-source-grounded benchmark for evaluating  Cross-Jurisdictional legal reasoning in LLMs across three jurisdictions covering China, California, and Germany. Built from authoritative legal sources, \textsc{CrossLex} aligns 55 legal issues spanning contract, consumer, criminal, family, and labor law, and constructs jurisdiction-aligned questions paired with answers and supporting citations. In total, \textsc{CrossLex} contains 6,149 instances organized into 385 fact groups, with all legal issues, answers, and cited authorities reviewed by legal professionals.
To disentangle basic legal knowledge from cross-jurisdictional reasoning, \textsc{CrossLex} defines three complementary tasks, including \textbf{single-jurisdiction reasoning (T1)}, \textbf{joint cross-jurisdictional comparison (T2)}, and \textbf{fine-grained cross-jurisdictional evaluation (T3)}. We further propose \textbf{Grounded Joint}, a metric that jointly assesses answer correctness and legal-source grounding, and provide a unified evaluation for streamlined benchmarking. Extensive experiments on representative LLMs show that, although current models can often answer legal questions correctly, they struggle to provide accurate cross-jurisdictional legal citations.
We hope that \textsc{CrossLex} will facilitate future research on source-grounded cross-jurisdictional legal reasoning.
\end{abstract}
\vspace{-1em}
\section{Introduction}

Large Language Models (LLMs) have made substantial progress on a range of legal NLP tasks~\citep{guha2023legalbench,fei2024lawbench,fan2025lexam}.
Existing legal benchmarks are typically organized around a single jurisdiction, a fixed body of law, or loosely connected collections of independent legal tasks
\citep{chalkidis2022lexglue,guha2023legalbench,fei2024lawbench}.
Consequently, they provide limited insight into a fundamental question in comparative legal reasoning, i.e., \textit{whether LLMs can derive jurisdiction-specific legal conclusions when the factual scenario remains unchanged but the legal system varies}.

\begin{figure}[!t]
  \centering
  \includegraphics[width=\columnwidth]{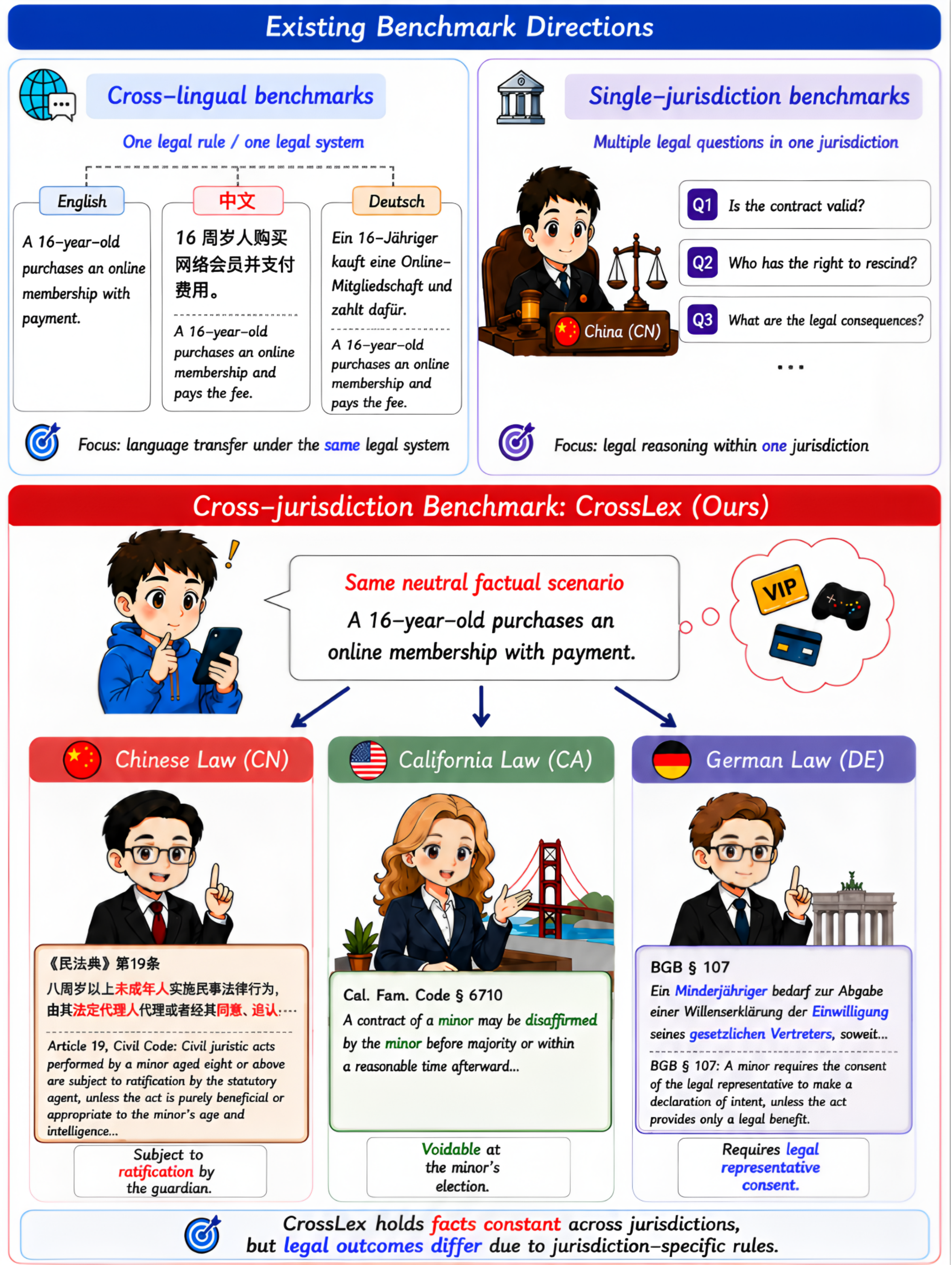}
    \vspace{-2em}
  \caption{
  The same factual scenario may lead to different legal conclusions across
  jurisdictions. \textsc{CrossLex} evaluates whether LLMs can distinguish
  jurisdiction-specific conclusions under neutral fact patterns.
  }
  \label{fig:crosslex_jurisdiction}
  \vspace{-2em}
\end{figure}

Legal reasoning is inherently jurisdiction-dependent, i.e., the same factual scenario may yield different legal conclusions under diverse legal systems. As illustrated in Figure~\ref{fig:crosslex_jurisdiction}, a 16-year-old's purchase of a paid online membership may have different contractual consequences across jurisdictions. For example, under Chinese law, validity may depend on ratification by the minor's guardian; under California law, the contract may be voidable at the minor's election; and under German law, it may require prior consent or subsequent approval from the minor's legal representative. It highlights a challenge for LLMs in cross-jurisdictional legal reasoning, where models must not only infer a plausible legal conclusion, but also associate it with the correct governing jurisdiction. Otherwise, they may overgeneralize familiar doctrines, conflate similar legal concepts, or incorrectly transfer rules across legal systems.
However, existing studies provide limited support for diagnosing jurisdictional confusion, as shown in Table~\ref{tab:benchmark_comparison}. Even when they cover multiple legal systems or languages, evaluation instances are typically considered in isolation, with each question asking for a conclusion under a single specified jurisdiction~\citep{chalkidis2021multieurlex,chalkidis2022lexglue,niklaus2024multilegalpile,fan2025lexam}.
They may assess single-jurisdiction legal knowledge, but they do not directly test controlled cross-jurisdictional comparison. Whether an LLM can reason over the same fact, distinguish divergent jurisdiction-specific conclusions, and correctly associate each conclusion with its governing legal.

\begin{table}[t]
\centering
\scriptsize
\setlength{\tabcolsep}{2.2pt}
\renewcommand{\arraystretch}{1.10}

\begin{tabularx}{\columnwidth}{
@{}
>{\raggedright\arraybackslash}X
>{\centering\arraybackslash}m{0.74cm}
>{\centering\arraybackslash}m{0.80cm}
>{\centering\arraybackslash}m{1.12cm}
>{\centering\arraybackslash}m{0.82cm}
>{\centering\arraybackslash}m{1.06cm}
@{}
}
\toprule

\textbf{Benchmark}
&
\shortstack{\textbf{Multi-}\\\textbf{Domain}}
&
\shortstack{\textbf{Multi-}\\\textbf{lingual}}
&
\shortstack{\textbf{Cross-}\\\textbf{Jurisdiction}}
&
\shortstack{\textbf{Aligned}\\\textbf{Facts}}
&
\shortstack{\textbf{Citation}\\\textbf{Grounding}}
\\

\midrule

COLIEE~\citeyearpar{goebel2023coliee}
& \pmark
& \xmark
& \pmark
& \xmark
& \pmark
\\

LexGLUE~\citeyearpar{chalkidis2022lexglue}
& \cmark
& \xmark
& \pmark
& \xmark
& \xmark
\\

LegalBench~\citeyearpar{guha2023legalbench}
& \cmark
& \xmark
& \xmark
& \xmark
& \xmark
\\

LawBench~\citeyearpar{fei2024lawbench}
& \cmark
& \xmark
& \xmark
& \xmark
& \pmark
\\

\midrule

MultiEURLEX~\citeyearpar{chalkidis2021multieurlex}
& \cmark
& \cmark
& \xmark
& \xmark
& \xmark
\\

LEXTREME~\citeyearpar{niklaus2023lextreme}
& \cmark
& \cmark
& \pmark
& \xmark
& \xmark
\\

LEXam~\citeyearpar{fan2025lexam}
& \cmark
& \cmark
& \pmark
& \xmark
& \xmark
\\

MultiLegalBench~\citeyearpar{ovcharov2026multilegalbench}
& \pmark
& \cmark
& \cmark
& \xmark
& \xmark
\\

\midrule

\textbf{\textsc{CrossLex}}
& \cmark
& \cmark
& \cmark
& \cmark
& \cmark
\\

\bottomrule
\end{tabularx}

\vspace{-1.0em}

\caption{
Comparison with representative legal benchmarks.
Multi-domain denotes coverage of multiple legal areas.
Cross-jurisdiction evaluates multiple legal systems, while aligned facts
requires the same neutral scenario across jurisdictions.
Citation grounding evaluates supporting legal sources.
\cmark{}, \xmark{}, and \pmark{} denote full, no, and partial support,
respectively.
}

\label{tab:benchmark_comparison}
\vspace{-1.3em}
\end{table}

\begin{figure*}[t]
\centering
\includegraphics[width=\textwidth]{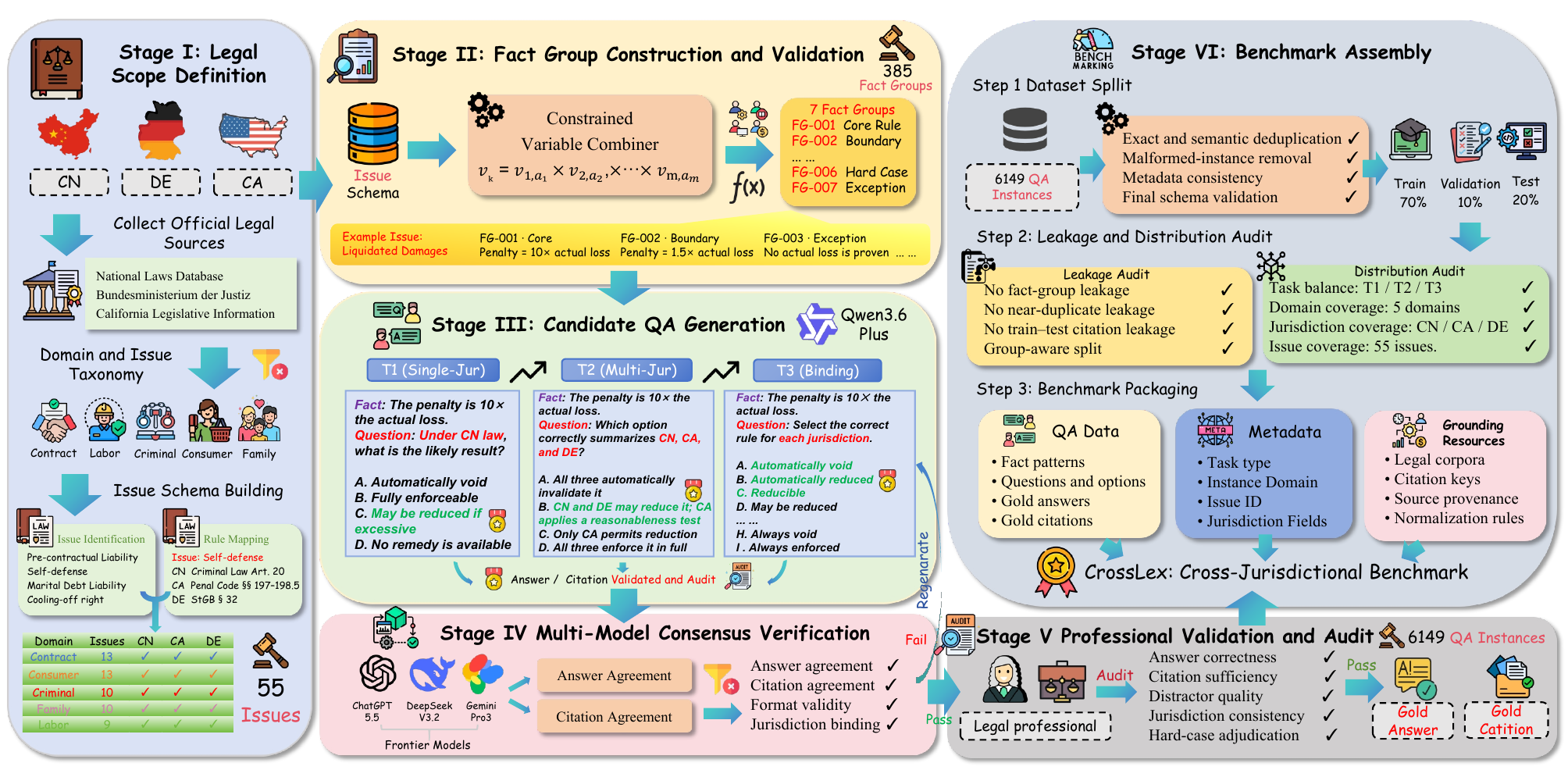}
\vspace{-1.5em}
\caption{Overview of the \textsc{CrossLex} construction framework. Neutral fact groups are aligned with jurisdiction-specific rules and source evidence, then instantiated into T1, T2, and T3 tasks.}
\vspace{-1.5em}
\label{fig:crosslex_framework}
\end{figure*}

To address this gap, we introduce \textbf{\textsc{CrossLex}}, a source-grounded cross-jurisdictional legal benchmark constructed through a six-step pipeline, as illustrated in Figure \ref{fig:crosslex_framework}. \textsc{CrossLex} evaluates LLMs on jurisdiction-aware legal reasoning over aligned, jurisdiction-neutral fact patterns. It spans three representative jurisdictions, including China, California (U.S.), and Germany, and five legal domains, such as contract law, consumer protection law, criminal law, family law, and labor law. Each fact pattern is written independently of any particular legal system and paired with jurisdiction-specific legal conclusions and supporting legal sources, enabling controlled comparison across legal systems under the same factual scenario.
\textsc{CrossLex} consists of three task formats with increasing difficulty. \textbf{T1: Single-Jurisdiction Judgment} evaluates whether a model can answer a legal question given a specified jurisdiction. \textbf{T2: Multi-Jurisdiction Joint Judgment} requires the model to select the correct combination of conclusions for the same fact pattern under different jurisdictions. \textbf{T3: Fine-Grained Multi-Jurisdiction Matching} asks the model to match each jurisdiction to its corresponding conclusion from a set of candidates. These tasks form a progression from jurisdiction-specific competence to joint cross-jurisdictional reasoning and fine-grained diagnosis of jurisdiction--conclusion alignment. We evaluate a diverse set of representative LLMs on \textsc{CrossLex}. Our results show that strong performance on T2 can conceal substantial failures on T3, i.e., models can often select the correct set of jurisdictional conclusions, but still fail to align each conclusion with its governing legal system. We further fine-tune LLMs at multiple parameter scales and show that adapted lightweight models can approach, and in some cases surpass, much larger proprietary models in source-grounded legal answering, particularly in citation grounding. 
Our contributions are threefold:
\begin{itemize}[leftmargin=*, itemsep=1pt, topsep=1pt, parsep=0pt]
    \item We introduce \textsc{CrossLex}, a source-grounded cross-jurisdictional legal benchmark spanning three jurisdictions, China, California (U.S.), and Germany, across five legal domains, paving the way for jurisdiction-aware legal reasoning.
    \item We design three increasingly challenging tasks over aligned, jurisdiction-neutral fact patterns, evaluating LLMs from single-jurisdiction judgment to multi-jurisdiction joint judgment and fine-grained jurisdiction--conclusion matching.
    \item We conduct extensive experiments with diverse LLMs, revealing persistent challenges in cross-jurisdictional legal reasoning. These findings highlight the need for further research on jurisdiction-aware legal models.

\end{itemize}

\vspace{-1em}
\section{\textsc{CrossLex}}
\label{sec:dataset_construction}

\subsection{Construction Overview}
\label{sec:construction_overview}

To construct \textsc{CrossLex}, we design an end-to-end benchmark construction pipeline, as illustrated in Figure~\ref{fig:crosslex_framework}. The pipeline comprises six stages: 
(1) defining the legal scope and cross-jurisdictional issue schemas; 
(2) constructing and validating controlled fact groups; 
(3) generating task-specific candidate QA instances; 
(4) verifying answer--citation consistency through multi-model consensus; 
(5) conducting professional legal validation and final auditing; and 
(6) assembling the validated instances into the final benchmark.

The pipeline begins with authoritative legal sources from China, California, and Germany. We identify functionally comparable legal issues across the three jurisdictions and encode each issue as a structured schema containing jurisdiction-specific rules, source evidence, controlled factual variables, distractor templates, and generation guardrails. Each issue schema is then expanded into seven controlled fact groups, which are instantiated into T1, T2, and T3 candidate questions. Candidate instances must pass both multi-model consensus checking and professional legal review before being included in the final benchmark.

\vspace{-1em}
\subsection{Stage I: Legal Scope and Issue Schema Definition}
\label{sec:stage1_issue_schema}

We define the legal scope of \textsc{CrossLex} over representative legal jurisdictions, i.e., China(CN), California (U.S. CA), and Germany(DE),  and five legal domains, such as contract law, consumer protection law, criminal law, family law, and labor law.
Let
\begin{equation}
    \mathcal{J}=\{\mathrm{CN},\mathrm{CA},\mathrm{DE}\}
\end{equation}
denote the set of jurisdictions considered in \textsc{CrossLex}, where
\(\mathrm{CN}\), \(\mathrm{CA}\), and \(\mathrm{DE}\) refer to China,
California (U.S.), and Germany, respectively.

\begin{figure*}[t]
    \centering
    \includegraphics[width=0.95\textwidth]{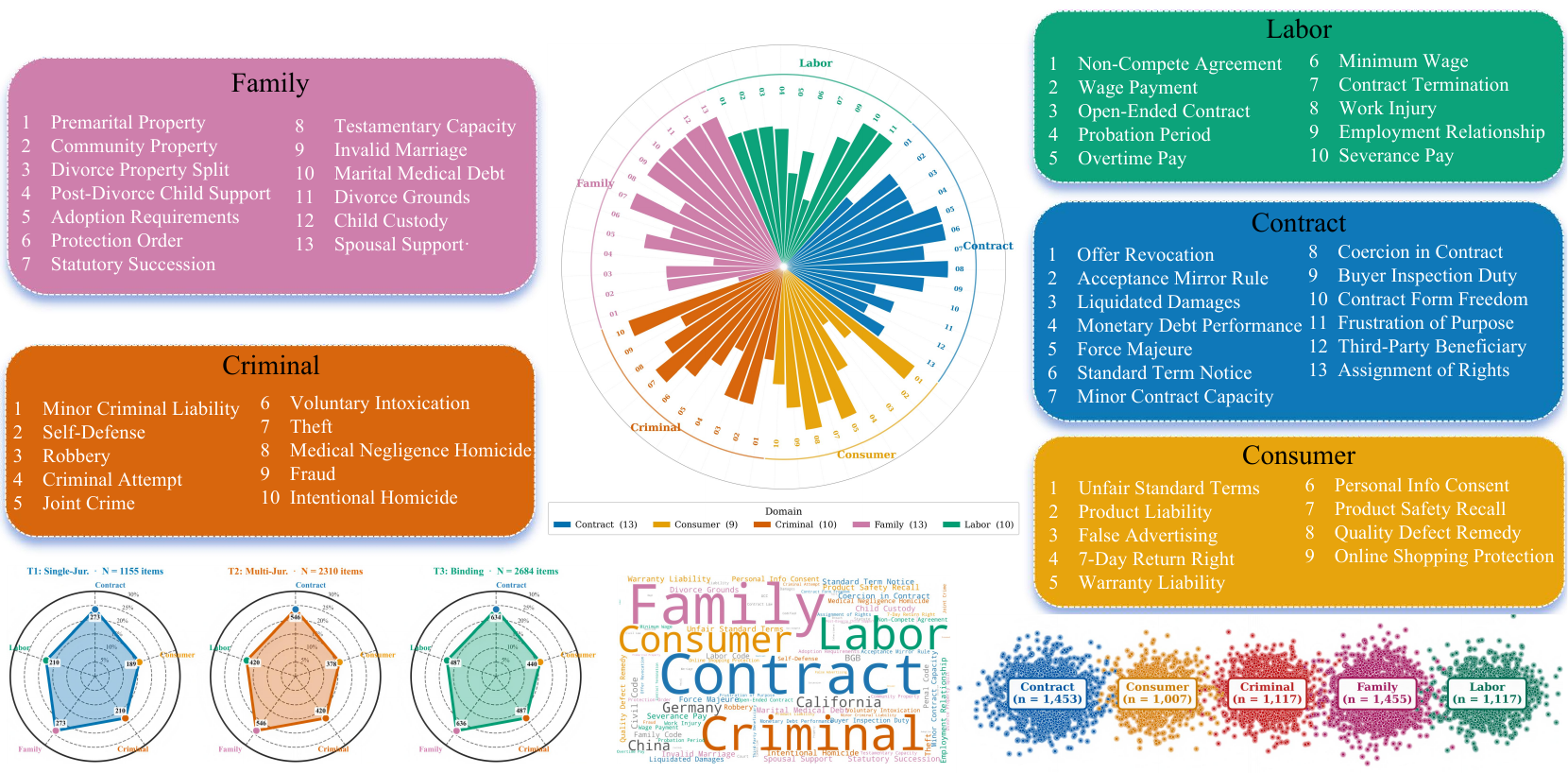}
    \vspace{-1.0em}
    \caption{
    Distribution of legal issues across the five domains in \textsc{CrossLex}.
    The central radial bar chart shows the number and relative distribution of issues in contract, consumer protection, criminal, family, and labor law, while the surrounding panels list the fine-grained legal issues included in each domain.
    }
    \label{fig:crosslex_domain_issue_distribution}
    \vspace{-1.5em}
\end{figure*}

For each domain, we identify functionally comparable legal issues that can
be meaningfully interpreted across all jurisdictions in \(\mathcal{J}\).
The resulting taxonomy contains 55 legal issues, as shown in
Figure~\ref{fig:crosslex_domain_issue_distribution}. Issue alignment is
functional rather than literal: although the corresponding rules may differ
in legal terminology, doctrinal structure, or legal consequences, they
address a shared legal problem.

We construct an issue schema for each issue \(i\).
\begin{equation}
\mathcal{S}_i =
\left(
R_{i,\mathrm{CN}},
R_{i,\mathrm{CA}},
R_{i,\mathrm{DE}},
F_i,
D_i,
B_i,
U_i
\right),
\end{equation}
where \(R_{i,j}\) denotes the jurisdiction-specific rule record for jurisdiction \(j \in \mathcal{J}\), \(F_i\) denotes the fact-specific slots, \(D_i\) denotes adversarial distractor templates, \(B_i\) denotes jurisdiction-specific forbidden terms and leakage constraints, and \(U_i\) denotes answer-uniqueness and issue-scope guardrails.

Each jurisdiction-specific rule record contains a concise rule summary, \(s_{i,j}\), an expected legal direction, \(y_{i,j}\), and supporting legal authorities, \(E_{i,j}\).
\begin{equation}
R_{i,j} = (s_{i,j}, y_{i,j}, E_{i,j}).
\end{equation}

To prevent models from relying on superficial pattern matching and to assess their ability to distinguish jurisdiction-specific legal rules, we introduce adversarial error templates at the issue-schema level. These templates define recurring failure modes, including jurisdiction confusion, rule swapping, overgeneralization, and incorrect legal citation, which guide later candidate option construction. For each issue schema, we define at least three jurisdiction-specific error templates covering these failure modes. We further use forbidden-term lists to prevent explicit jurisdictional leakage and apply uniqueness guardrails to ensure that each multiple-choice item has exactly one legally defensible answer. Detailed distractor examples and generation prompts are included in the supplementary materials.

\vspace{-1em}
\subsection{Stage II: Fact Group Construction and Validation}
\label{sec:stage2_fact_groups}

Each locked issue schema is expanded into multiple controlled fact groups.
For issue \(i\), let
\[
F_i=\{V_{i,\ell}\}_{\ell=1}^{m_i}
\]
denote its \(m_i\) controllable factual slots, where \(V_{i,\ell}\)
is the set of candidate values for slot \(\ell\).
The \(k\)-th valid variable combination is
\begin{equation}
\begin{aligned}
\mathbf{v}_{i,k}
&=
\left(
v_{i,1,a^{(k)}_1},
\ldots,
v_{i,m_i,a^{(k)}_{m_i}}
\right), \\
&\text{s.t.}\quad
v_{i,\ell,a^{(k)}_\ell}\in V_{i,\ell},
\qquad
\mathbf{v}_{i,k}\models\mathcal{G}_i,
\end{aligned}
\end{equation}
where \(a^{(k)}_\ell\) is the index of the value selected for slot
\(\ell\), and
\(\mathbf{v}_{i,k}\models\mathcal{G}_i\) indicates that the combination
satisfies all issue-specific compatibility, legality, and uniqueness
constraints.

Each valid combination is instantiated into a fact group \(g_{i,k}\),
where \(k\in\{1,\ldots,K\}\) and \(K=7\). Its jurisdiction-specific
answer directions are represented as
\begin{equation}
Y_{i,k}
=
\left(
y_{i,k,\mathrm{CN}},
y_{i,k,\mathrm{CA}},
y_{i,k,\mathrm{DE}}
\right).
\end{equation}
Here, \(y_{i,k,j}\) denotes the final legal direction for jurisdiction
\(j\) after instantiating the factual variables of \(g_{i,k}\), whereas
\(y_{i,j}\) denotes the issue-level default direction defined in Stage I.

Most fact groups preserve the issue-level rule direction, while selected
divergence-focused groups intentionally produce different outcomes across
jurisdictions. These groups are particularly important for constructing
T3 jurisdiction-binding instances. Before locking, each fact group is
audited for fact specificity, constraint completeness, issue consistency,
cross-issue risk, and human-review requirements. Failed groups are
regenerated or revised. This process yields \(385=55\times7\) validated
fact groups from 55 issue schemas.

\vspace{-1em}
\subsection{Stage III: Candidate QA Generation}
\label{sec:stage3_candidate_generation}

Each validated fact group is transformed into three task-specific QA
formats. Let
\(\mathcal{T}=\{\mathrm{T1},\mathrm{T2},\mathrm{T3}\}\)
denote the task set. For each \(g_{i,k}\) and \(t\in\mathcal{T}\),
\begin{equation}
\mathcal{Q}^{\mathrm{cand}}_{i,k,t}
=
G_\theta
\left(
f_{i,k},
Y_{i,k},
\mathcal{S}_i,
t
\right),
\end{equation}
where \(\mathcal{S}_i\) is the corresponding issue schema and \(G_{\theta}\) denotes the candidate-generation LLM. In \textsc{CrossLex}, we use Qwen-3.6-Plus to generate task-specific questions, answer options,
and distractors from the validated fact groups, while all legal directions and final annotations are determined through source-grounded validation.

We design three task formats with increasing difficulty, as shown in Stage III of Figure \ref{fig:crosslex_domain_issue_distribution}.
For T1, the generator produces a single-jurisdiction multiple-choice question for one specified jurisdiction. For T2, it constructs a joint multiple-choice item whose correct option encodes the complete CN--CA--DE conclusion tuple. For T3, it produces a candidate-conclusion pool and requires separate jurisdiction-to-conclusion assignments.
Correct answer candidates are instantiated from the expected answer directions. During candidate QA generation, the predefined adversarial error templates are instantiated into concrete distractor options. These distractors simulate realistic legal reasoning conditions, where similar doctrines, related legal issues, and cross-jurisdictional rules may lead to plausible but incorrect conclusions. They prevent models from relying on superficial patterns and enable evaluation of fine-grained jurisdiction-aware reasoning.
Distractors are generated from the locked adversarial templates in the issue schema. They cover direction reversal, jurisdiction-rule swapping, overgeneralization, and citation-related confusion. Candidate-level guardrails remove items containing forbidden jurisdictional terms, duplicated options, multiple defensible answers, or unintended adjacent legal issues.

\subsection{Stage IV: Multi-Model Consensus Verification}
\label{sec:stage4_multimodel}

Candidate QA instances are independently evaluated by three LLM validators:
GPT-5.5~\citep{openai2026gpt55},
DeepSeek-V3.2~\citep{deepseekai2025v32}, and
Gemini~3~Pro~\citep{googledeepmind2026gemini3pro}. Each validator receives the same fact pattern, question, candidate answers, and task instructions, but does not observe the provisional gold label from Stage III. Each model independently predicts the answer and provides supporting legal authorities for all applicable jurisdictions. Unlike conventional answer-only validation, our validation process jointly examines answer correctness and legal-source grounding: a candidate is retained only when the predicted conclusion is consistent and the cited authorities match the corresponding jurisdiction-specific evidence after citation normalization.

For each candidate \(x\in\mathcal{X}^{\mathrm{cand}}\), three validators
\(V^{(m)}\), where \(m\in\{1,2,3\}\), independently produce
\begin{equation}
o_x^{(m)}
=
V^{(m)}(x)
=
\left(
\hat{y}_x^{(m)},
\{\hat{E}_{x,j}^{(m)}\}_{j\in\mathcal{J}_x}
\right),
\end{equation}
where \(\mathcal{J}_x\subseteq\mathcal{J}\) is the set of jurisdictions
involved in item \(x\), \(\hat{y}_x^{(m)}\) is the predicted answer, and
\(\hat{E}_{x,j}^{(m)}\) is the supporting legal evidence predicted for
jurisdiction \(j\).
Raw citations are normalized into canonical citation identifiers before comparison. An item passes model verification only when all validators agree on the answer and their normalized citations are consistent with the allowed source evidence.
\begin{equation}
C_{\mathrm{model}}(x)
=
C_{\mathrm{ans}}(x)
\land
C_{\mathrm{cite}}(x)
\land
C_{\mathrm{fmt}}(x)
\land
C_{\mathrm{jur}}(x),
\end{equation}
where the four terms denote answer agreement, citation agreement, format validity, and jurisdiction consistency. 
Detailed validator prompts, output schemas, and verification criteria are provided in the supplementary materials.
Items that fail answer or citation consensus are rejected or returned to the candidate-generation stage. Items that pass are treated as model-verified QA candidates and forwarded to professional validation.

\vspace{-1em}
\subsection{Stage V: Professional Validation and Final Audit}
\label{sec:stage5_professional}

Model consensus provides a large-scale consistency check over both answers and cited legal authorities. 
Since agreement among LLMs does not guarantee legal validity, all consensus-passed candidates are further reviewed by legal professionals for final verification.
Legal reviewers with formal legal training participated in the final validation stage. 
Each item was independently reviewed by two legally trained annotators.
Reviewers were assigned according to jurisdictional expertise, and
disagreements were resolved through adjudication with reference to the
official legal sources.
They manually audit each item along five dimensions:
\begin{itemize}[leftmargin=*, itemsep=1pt, topsep=1pt]
    \item \textbf{Answer correctness}: whether the selected option or jurisdiction mapping follows from the relevant law;
    \item \textbf{Citation sufficiency}: whether the cited authority directly supports the legal conclusion;
    \item \textbf{Distractor quality}: whether distractors are plausible but legally distinguishable;
    \item \textbf{Jurisdiction consistency}: whether each rule and citation is associated with the correct legal system;
    \item \textbf{Hard-case adjudication}: whether ambiguous or borderline instances require revision or exclusion.
\end{itemize}

Reviewers may accept, revise, or reject an item. Revised items are returned for revalidation. Only items that pass the professional audit receive final gold answers and gold citations. This process produces 6,149 expert-validated QA instances. Detailed review guidelines and annotation instructions are provided in the supplementary materials.

\subsection{Stage VI: Benchmark Assembly}
\label{sec:stage6_assembly}

The 6,149 expert-validated QA instances are curated and assembled into the final \textsc{CrossLex} benchmark. We first perform exact and semantic deduplication, remove malformed instances, verify metadata consistency, and conduct final schema validation.
To prevent leakage, all QA instances derived from the same fact group are assigned to the same split. In particular, T1, T2, and T3 variants that share the same underlying fact group cannot be separated across training and evaluation splits. We further audit near-duplicate questions and ensure that retrieval resources and query-expansion lexicons are constructed without using development or test information.
The data is partitioned into training, validation, and test sets using a 70\%/10\%/20\% split. We audit the distributions of task types, legal domains, jurisdictions, and issue coverage across the three splits.
The final benchmark contains three components: 
\textbf{QA data}: neutral fact patterns, questions, options, gold answers, and gold citations;
\textbf{Metadata}: task type, domain, issue identifier, fact-group identifier, jurisdiction fields, and split information;
\textbf{Grounding resources}: legal corpora, normalized citation keys, source provenance, and citation-normalization rules.

\begin{table*}[t]
\centering
\small
\setlength{\tabcolsep}{1.2pt}
\renewcommand{\arraystretch}{1.02}

\begin{tabular*}{\textwidth}{@{\extracolsep{\fill}}l
*{6}{c}
*{3}{c}
*{6}{c}
c@{}}

\toprule

\multirow{2}{*}{\textbf{Model}}
&
\multicolumn{6}{c}{\textbf{T1}}
&
\multicolumn{3}{c}{\textbf{T2}}
&
\multicolumn{6}{c}{\textbf{T3}}
&
\multirow{2}{*}{\textbf{GJ-Avg}}
\\

\cmidrule(lr){2-7}
\cmidrule(lr){8-10}
\cmidrule(lr){11-16}

&
\textbf{All}
&
\textbf{CN}
&
\textbf{CA}
&
\textbf{DE}
&
\textbf{Citation-F1}
&
\textbf{GJ}
&
\textbf{Joint}
&
\textbf{Citation-F1 }
&
\textbf{GJ}
&
\textbf{All}
&
\textbf{CN}
&
\textbf{CA}
&
\textbf{DE}
&
\textbf{Citation-F1 }
&
\textbf{GJ}
&
\\

\midrule

\multicolumn{17}{c}{\textit{Closed-source LLMs}}
\\
\midrule

GPT-5.6
&99.6
&\textbf{100.0}
&\textbf{100.0}
&98.7
&60.0
&\textbf{59.8}
&98.1
&39.3
&38.6
&43.7
&47.6
&46.6
&50.7
&18.7
&16.0
&38.2
\\

Claude-4.8
&98.3
&98.7
&98.7
&97.4
&\textbf{60.6}
&59.6
&98.7
&\textbf{52.1}
&\textbf{51.5}
&\textbf{86.9}
&\textbf{91.8}
&\textbf{95.3}
&\textbf{95.5}
&\textbf{51.5}
&\textbf{45.2}
&\textbf{52.1}
\\

Gemini-3.1
&\textbf{100.0}
&\textbf{100.0}
&\textbf{100.0}
&\textbf{100.0}
&46.1
&46.1
&\textbf{99.1}
&49.7
&49.5
&40.5
&45.3
&42.4
&45.3
&24.4
&21.5
&39.0
\\

Qwen-3.7
&\textbf{97.8}
&\textbf{100.0}
&\textbf{100.0}
&93.5
&60.6
&59.1
&\textbf{99.6}
&\textbf{49.8}
&\textbf{49.6}
&\textbf{71.8}
&\textbf{79.5}
&\textbf{89.9}
&\textbf{94.4}
&48.2
&\textbf{34.8}
&47.8
\\

\midrule

\multicolumn{17}{c}{\textit{Open-source Large LLMs}}
\\
\midrule

GLM-5.2
&\textbf{97.8}
&98.7
&\textbf{100.0}
&\textbf{94.8}
&\textbf{67.9}
&\textbf{66.6}
&99.1
&46.5
&46.2
&62.7
&78.0
&88.1
&82.6
&\textbf{50.1}
&32.1
&\textbf{48.3}
\\

DS-V4
&84.8
&96.1
&85.7
&72.7
&55.0
&48.5
&86.6
&41.2
&39.6
&30.4
&39.4
&45.5
&47.9
&27.4
&14.3
&34.1
\\

Llama-3.3-70B
&\textbf{96.1}
&94.8
&\textbf{97.4}
&\textbf{96.1}
&52.2
&50.4
&97.0
&29.7
&29.3
&27.6
&42.4
&77.8
&66.0
&25.8
&8.0
&29.3
\\

\midrule

\multicolumn{17}{c}{\textit{Open-source Medium LLMs}}
\\
\midrule

Qwen3.5-35B
&94.4
&\textbf{98.7}
&\textbf{97.4}
&87.0
&58.7
&56.0
&98.3
&\textbf{49.8}
&\textbf{49.3}
&37.3
&56.7
&75.4
&70.5
&45.5
&16.2
&40.5
\\

Qwen3.5-27B
&\textbf{96.1}
&\textbf{98.7}
&\textbf{97.4}
&92.2
&\textbf{63.1}
&\textbf{61.0}
&\textbf{99.4}
&49.2
&49.1
&\textbf{51.3}
&\textbf{65.7}
&\textbf{84.7}
&\textbf{82.8}
&\textbf{46.4}
&\textbf{24.7}
&\textbf{44.9}
\\

DS-R1-14B
&87.4
&93.5
&96.1
&72.7
&48.0
&46.2
&93.9
&25.0
&23.8
&16.0
&47.9
&60.4
&44.0
&19.6
&3.5
&24.5
\\

Qwen3-14B
&82.3
&94.8
&90.9
&61.0
&52.9
&46.7
&90.7
&32.0
&30.8
&21.8
&42.7
&58.2
&43.8
&21.7
&5.1
&27.5
\\

\midrule

\multicolumn{17}{c}{\textit{Open-source Small LLMs}}
\\
\midrule

GLM-4-9B
&78.8
&85.7
&88.3
&62.3
&\textbf{53.5}
&46.4
&85.9
&27.0
&24.6
&5.2
&31.0
&38.1
&30.4
&22.8
&1.0
&24.0
\\

DS-R1-8B
&84.0
&\textbf{98.7}
&87.0
&66.2
&45.7
&43.6
&89.4
&21.7
&20.5
&17.7
&50.7
&\textbf{64.9}
&43.8
&15.6
&3.2
&22.4
\\

Qwen3-8B
&80.5
&94.8
&89.6
&57.1
&50.3
&45.4
&89.2
&24.6
&22.7
&12.1
&37.9
&41.2
&34.3
&18.7
&2.3
&23.4
\\

Qwen3.5-9B
&\textbf{92.5}
&98.4
&\textbf{94.7}
&\textbf{91.8}
&37.2
&36.6
&\textbf{94.6}
&25.8
&25.1
&15.5
&\textbf{60.1}
&47.6
&\textbf{43.9}
&\textbf{23.9}
&3.7
&21.8
\\

Qwen3.5-4B
&77.4
&98.4
&87.3
&64.8
&37.1
&36.0
&79.0
&17.4
&16.1
&10.6
&51.6
&44.0
&42.1
&17.9
&2.3
&18.1
\\

Qwen3.5-2B
&46.2
&89.3
&53.7
&37.8
&29.1
&24.4
&70.7
&17.4
&13.7
&1.2
&36.0
&28.0
&26.1
&9.6
&0.0
&12.7
\\

\bottomrule

\end{tabular*}
\vspace{-1em}
\caption{
Main zero-shot results of diverse LLMs on the test set of
\textsc{CrossLex} (\%). Apart from Citation-F1 and GJ, other columns are Accuracy, covering three jurisdictions (CN, CA, DE) across three tasks (T1, T2, T3).T3 All denotes exact jurisdiction-binding accuracy, and GJ-Avg is the macro-average of the T1, T2, and T3 GJ scores.
Best results within each model group are shown in \textbf{bold}.
}
\label{tab:main_results_crosslex}
\vspace{-1.5em}
\end{table*}
\vspace{-1em}
\section{Experiments}
\label{sec:experiments}

\subsection{Dataset Statistics}
\label{sec:dataset_statistics}
\textsc{CrossLex} contains 55 aligned legal issues, 385 fact groups, and
6,149 professionally validated QA instances across five domains and three
jurisdictions. Each instance includes a gold answer, jurisdiction-specific
citations, normalized citation identifiers, task metadata, and source
provenance.
\vspace{-1em}
\subsection{Experimental Setup}
\label{sec:experimental_setup}

We evaluate \textsc{CrossLex} under a progressive setting: T1 evaluates
single-jurisdiction judgment, T2 requires joint prediction of the
CN--CA--DE conclusion tuple under aligned facts, and T3 evaluates
jurisdiction--conclusion binding.
\vspace{-0.5em}
\subsection{Baselines}
\vspace{-0.3em}
\paragraph{Representative LLMs}
We evaluate representative LLMs across different scales and access
settings. Closed-source and API-accessible models include
GPT-5.6~\citep{openai2026gpt56},
Claude-Opus-4.8~\citep{anthropic2026claudeopus48},
Gemini-3.1-Pro~\citep{googledeepmind2026gemini31pro}, and
Qwen3.7-Max~\citep{alibaba2026qwen37max}.
Large open-weight models include
DeepSeek-V4-Pro~\citep{deepseek2026v4},
GLM-5.2~\citep{glm5team2026glm5}, and
Llama-3.3-70B-Instruct~\citep{meta2024llama33}.
Medium-scale models include
Qwen3.5-35B-A3B and Qwen3.5-27B~\citep{qwen2026qwen35},
and DeepSeek-R1-Distill-Qwen-14B~\citep{deepseekai2025deepseekr1}.
Small-scale models include
GLM-4-9B~\citep{glmteam2024chatglm},
DeepSeek-R1-Distill-8B~\citep{deepseekai2025deepseekr1},
Qwen3-14B and Qwen3-8B~\citep{yang2025qwen3},
and Qwen3.5-2B, Qwen3.5-4B, and Qwen3.5-9B
\citep{qwen2026qwen35}.
\vspace{-0.5em}
\paragraph{Multilingual BM25-RAG}
\label{sec:multilingual_bm25}
Because \textsc{CrossLex} involves heterogeneous languages and citation
conventions across Chinese, California, and German law, conventional
monolingual retrieval is not well suited to retrieving relevant legal
authorities across jurisdictions. We therefore construct a multilingual
BM25-based RAG baseline by combining BM25 retrieval
\citep{robertson2009bm25} with retrieval-augmented generation
\citep{lewis2020retrieval}, and evaluate whether external legal evidence
improves source-grounded reasoning.
Given a legal query, the retriever first identifies the target jurisdiction and retrieves relevant authorities from the corresponding legal corpus.
The retrieved evidence is then provided to the model for answer generation.
Details of these models are provided in the supplementary materials.

\begin{figure*}[t]
\centering
\includegraphics[width=\textwidth]{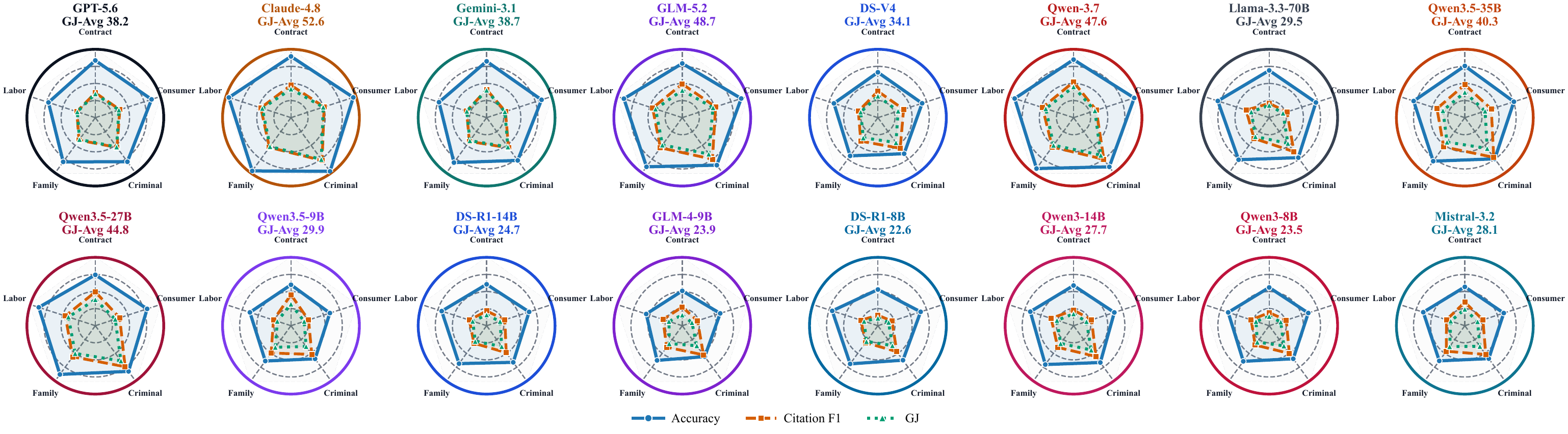}
\vspace{-2em}
\caption{
Per-model multi-metric radar profiles on \textsc{CrossLex}
(T1/T2/T3 Accuracy, Citation-F1, and GJ) across five law domains (Contract, Consumer protection, Criminal, Family, and Labor law).
Each panel shows one model under unified exact-set scoring with multiple denominators.
}
\vspace{-1.5em}
\label{fig:main_radar_each_model}
\end{figure*}

\vspace{-0.5em}
\subsection{Evaluation Metrics}
\label{sec:evaluation_metrics}

We report answer correctness using Accuracy and citation quality using
Citation-F1. Citation-F1 is computed over normalized legal-source
identifiers independently of answer correctness.
For T1, we measure whether the model selects the correct legal conclusion under the specified jurisdiction.
For T2, Joint Accuracy requires the complete CN--CA--DE conclusion tuple to be correct.
For T3, Jurisdiction Binding Accuracy measures whether each predicted conclusion is assigned to the correct jurisdiction.
Since a legally grounded response requires both a correct conclusion and
appropriate supporting authority, we define \textbf{Grounded Joint (GJ)} as
\begin{equation}
\mathrm{GJ}
=
\frac{1}{N}
\sum_{n=1}^{N}
\mathbf{1}
\left[
\hat{y}_n=y_n
\right]
\cdot
\mathrm{F1}^{\mathrm{cite}}_n,
\end{equation}
where \(N\) is the number of evaluated instances,
\(\hat{y}_n\) and \(y_n\) are the predicted and gold answers, and
\(\mathrm{F1}^{\mathrm{cite}}_n\) is the Citation-F1 score between the
predicted and gold normalized citation sets for instance \(n\).

\begin{figure}[t]
\centering
\includegraphics[width=\columnwidth]{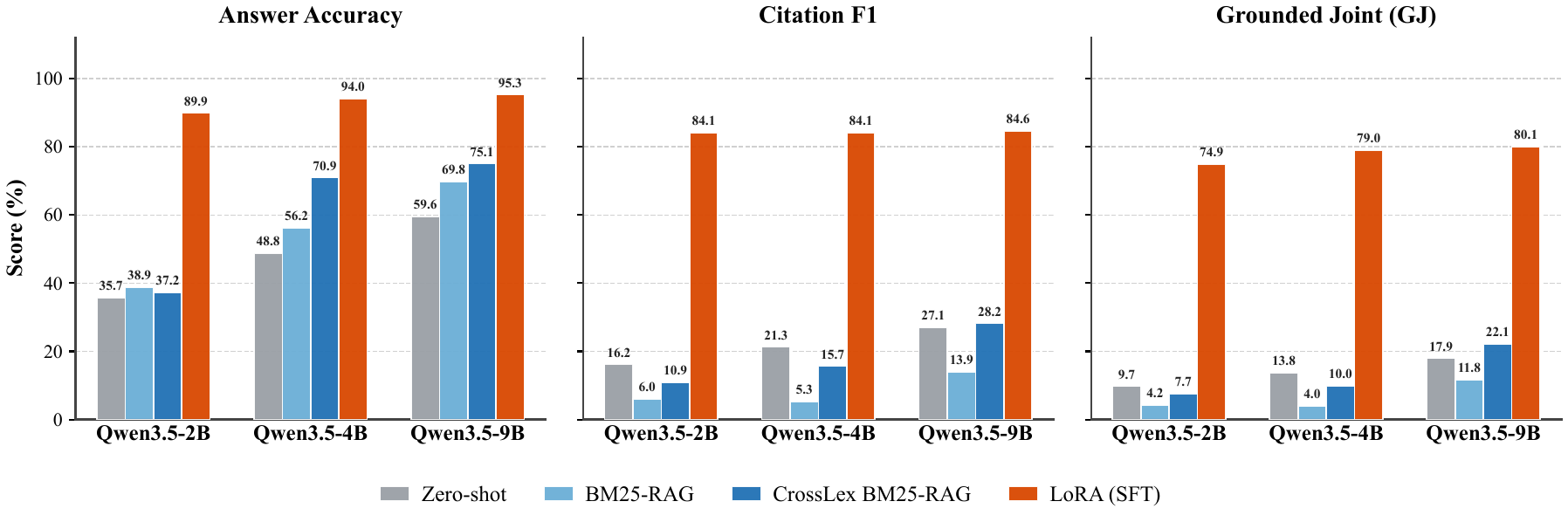}
\vspace{-2em}
\caption{
Adaptation results on \textsc{CrossLex} across model scales.
}
\vspace{-1.5em}
\label{fig:lora_scale_adaptation}
\end{figure}

\subsection{Main Results}
\label{sec:main_results}

Table~\ref{tab:main_results_crosslex} reports the overall results on
\textsc{CrossLex}. Performance generally improves with model scale, and
larger models tend to achieve stronger overall results. However, even the
best-performing models remain unreliable in cross-jurisdictional legal
citation grounding.

Models perform strongly on T1, suggesting that current LLMs possess
substantial single-jurisdiction legal knowledge. This capability, however,
does not fully transfer to cross-jurisdictional reasoning. In particular,
T3 reveals a pronounced \textbf{binding gap}: models may identify plausible
legal conclusions under the same fact pattern, yet fail to associate each
conclusion with the correct governing jurisdiction. This indicates that
recognizing applicable legal rules and preserving jurisdiction-specific
boundaries are distinct capabilities.

Grounded Joint (GJ) scores are also consistently lower than answer
accuracy, exposing a substantial \textbf{grounding gap}. Models may select
the correct answer while providing unsupported, incomplete, or
jurisdiction-inconsistent citations. As shown in
Fig.~\ref{fig:main_radar_each_model}, citation grounding further varies
across CN, CA, and DE, as well as across legal domains. Models tend to
struggle more in domains with stronger jurisdictional variation, reflecting
differences in legal language, citation conventions, and source structures.
Overall, these findings show that legal model evaluation should move beyond
answer accuracy toward jurisdiction-aware and source-grounded assessment.
\vspace{-0.5em}
\subsection{CrossLex-guided Legal Adaptation}
\label{sec:crosslex_adaptation}

We examine whether retrieval-augmented generation (RAG) and
\textsc{CrossLex} supervision improve source-grounded legal reasoning,
as reported in Fig.~\ref{fig:lora_scale_adaptation}. Compared with
zero-shot inference, multilingual BM25-RAG supplies additional legal
evidence and improves answer selection in some settings. Nevertheless,
its gains in GJ remain limited, indicating that retrieval alone is
insufficient for jurisdiction-aware legal reasoning. Models must also
associate retrieved authorities with the correct jurisdiction-specific
conclusions.

We then fine-tune lightweight models on the \textsc{CrossLex} training
set using Low-Rank Adaptation (LoRA)~\cite{hu2022lora}. LoRA
substantially improves citation quality and Grounded Joint scores while
maintaining high answer accuracy. These results suggest that
\textsc{CrossLex} is useful not only for evaluation, but also as
supervision for learning jurisdiction-aware and source-grounded legal
reasoning.
\vspace{-1.0em}
\section{Related Works}

\subsection{Legal Benchmarks}

Existing legal benchmarks cover judgment prediction, question answering,
case holding identification, contract review, statutory reasoning, legal
retrieval, and general legal reasoning. Representative examples include
CAIL for Chinese legal judgment prediction, JEC-QA for judicial-exam
question answering, CaseHOLD for U.S. case holding identification, CUAD for
contract review, and recent LLM-oriented benchmarks such as LegalBench,
LawBench, and LEXam~\citep{xiao2018cail,zhong2020jecqa,
zheng2021casehold,hendrycks2021cuad,guha2023legalbench,
fei2024lawbench,fan2025lexam}.

Another line of work studies multilingual and multi-system legal
understanding through benchmarks such as MultiEURLEX, LexGLUE, LEXTREME,
and MultiLegalBench~\citep{chalkidis2021multieurlex,
chalkidis2022lexglue,niklaus2023lextreme,ovcharov2026multilegalbench},
as well as multilingual legal corpora and pretrained models such as
MultiLegalPile, LEGAL-BERT, and Lawformer~\citep{
niklaus2024multilegalpile,chalkidis2020legalbert,xiao2021lawformer}.

These resources are valuable, but they generally evaluate jurisdictions,
languages, or tasks independently. Even when multiple legal systems or
languages are covered, they do not align identical factual scenarios across
jurisdictions or test whether models bind each legal conclusion to the
correct legal system.
In contrast, \textsc{CrossLex} focuses on cross-jurisdictional reasoning
under aligned neutral fact patterns, where the same facts are paired with
jurisdiction-specific rules and legal outcomes. 
\vspace{-0.8em}
\subsection{Legal Reasoning}
Legal reasoning differs from general question answering because legal conclusions should be supported by authoritative statutes, cases, or regulatory sources. Prior work has therefore studied legal retrieval, legal entailment, contract clause extraction, and citation-sensitive evaluation
\citep{hendrycks2021cuad,goebel2023coliee,lewis2020retrieval,karpukhin2020dense,izacard2021leveraging}.
Retrieval-augmented generation and dense retrieval methods are widely used for knowledge-intensive tasks, while legal retrieval benchmarks such as COLIEE emphasize the need to connect legal questions with relevant legal texts. However, source grounding alone does not resolve the cross-jurisdictional problem. i.e., a model may retrieve or cite plausible legal sources but still assign the resulting conclusion to the wrong legal system. \textsc{CrossLex} therefore pairs each jurisdiction-specific conclusion with supporting legal evidence and further evaluates fine-grained jurisdiction-to-conclusion matching, making it possible to diagnose whether models preserve legal-system boundaries under aligned facts.

\vspace{-0.8em}
\section{Conclusion}

We introduce \textsc{CrossLex}, a source-grounded benchmark for
cross-jurisdictional legal reasoning over aligned neutral facts from
Chinese, California, and German law across five domains.
It defines three tasks covering single-jurisdiction judgment (T1),
joint cross-jurisdictional comparison (T2), and fine-grained
jurisdiction--conclusion binding (T3).
Experiments show that frontier models approach ceiling performance on T2,
while T3 reveals substantial failures in assigning legal conclusions to
the correct jurisdiction, highlighting the limitations of conventional
single-jurisdiction and multilingual legal QA evaluation.

\vspace{-0.8em}
\section*{Ethics Statement}

This work involves human participants serving as legal reviewers.
They review synthesized neutral fact patterns and publicly available legal
sources and do not provide private or personally identifiable data.
\textsc{CrossLex} is intended for research evaluation only and should not
replace professional legal judgment. Real-world use should involve qualified
legal professionals and appropriate human oversight.

\bibliography{aaai2027}
\end{document}